\documentclass[10pt,twocolumn,letterpaper]{article}

\usepackage[pagenumbers]{cvpr}      

\usepackage{bbding}
\usepackage{graphicx}
\usepackage{amsmath}
\usepackage{amssymb}
\usepackage{booktabs}

\usepackage[pagebackref,breaklinks,colorlinks]{hyperref}

\usepackage[capitalize]{cleveref}
\crefname{section}{Sec.}{Secs.}
\Crefname{section}{Section}{Sections}
\Crefname{table}{Table}{Tables}
\crefname{table}{Tab.}{Tabs.}

\def\cvprPaperID{*****} 
\def\confName{CVPR}
\def\confYear{2023}

\begin{document}


\title{Group DETR v2: Strong Object Detector with Encoder-Decoder Pretraining}
\author{
Qiang Chen$^{1*}$, Jian Wang$^{1}$\thanks{Equal contribution.}~~, Chuchu Han$^{1*}$, Shan Zhang$^2$, Zexian Li$^3$, Xiaokang Chen$^4$, Jiahui Chen$^3$ \\ Xiaodi Wang$^1$, Shuming Han$^1$, Gang Zhang$^1$, Haocheng Feng$^1$, Kun Yao$^1$, Junyu Han$^1$, Errui Ding$^1$\\ Jingdong Wang$^1$\thanks{Corresponding author.} \\
\\
$^1$Baidu VIS~~~~$^2$Australian National University~~~~$^3$Beihang University~~~~$^4$Peking University\\
}

\maketitle

\begin{abstract} 
We present a strong object detector with 
encoder-decoder pretraining and finetuning. 
Our method, called Group DETR v2, is built upon a vision transformer encoder ViT-Huge~\cite{dosovitskiy2020image}, 
a DETR variant DINO~\cite{zhang2022dino}, and an efficient DETR training method 
Group DETR~\cite{chen2022group}. 
The training process consists of self-supervised pretraining and finetuning a ViT-Huge encoder on ImageNet-1K, 
pretraining the detector on Object365, 
and finally finetuning it on COCO. 
Group DETR v2 achieves $\textbf{64.5}$ mAP on COCO test-dev, and
establishes a new SoTA on the COCO leaderboard\footnote{\url{https://paperswithcode.com/sota/object-detection-on-coco}}.
\end{abstract}

\section{Introduction} \label{sec:intro}

Recent studies show the effectiveness of transformer models at scale~\cite{dosovitskiy2020image,zhai2022scaling,liu2022swin}.
With encoder pretraining on large-scale data~\cite{deng2009imagenet,ridnik2021imagenet}, the models~\cite{bao2021beit,he2022masked,xie2022simmim,chen2022context,peng2022beit,wang2022image} are able to achieve superior results 
on various vision tasks, including object detection. 
With supervised encoder-decoder pretraining on a large-scale dataset, Object365~\cite{shao2019objects365}, DINO~\cite{zhang2022dino} achieves a state-of-the-art result on COCO~\cite{lin2014microsoft}.

Our method, {\em Group DETR v2}, is built upon ViT-Huge, DINO, and Group DETR. 
We adopt an encoder-decoder pretraining and finetuning pipeline: 
pretraining and then
finetuning a ViT-Huge encoder on ImageNet-1K~\cite{deng2009imagenet}, 
pretraining the detector, both the encoder and the decoder, on Object365, 
and finally finetuning it on COCO. 
Group DETR v2 achieves $\textbf{64.5}$ mAP on COCO test-dev~\cite{lin2014microsoft} (Table~\ref{tab:leaderboard} and Table~\ref{tab:coco_dev}), setting a new record for COCO object detection. 
We expect that the results will be further improved with more data and larger models. 

\section{Method} \label{sec:method}

\subsection{Architecture} \label{modeling}

\paragraph{Encoder.}
We adopt a ViT-Huge as the encoder.  We resort to the self-supervised pretrained model, Vit-Huge, e.g., with the MIM method CAE~\cite{chen2022context},
which shows superior performance on downstream tasks. 
We follow ViTDet~\cite{li2022exploring} to build multi-scale feature maps for multi-scale DETR.

\paragraph{Decoder.}
We adopt the transformer encoder-decoder framework as the decoder that shows promising detection results, including DETR~\cite{carion2020end}, Conditional DETR~\cite{meng2021conditional}, DAB-DETR~\cite{liu2022dab}, Deformable DETR~\cite{zhu2020deformable}, DN-DETR~\cite{li2022dn}, and DINO\cite{zhang2022dino}. Group DETR~\cite{chen2022group} provides further progress in improving the training convergence speed and the detection performance of various DETR variants. 
We build our detection decoder by combining DINO~\cite{zhang2022dino} 
into Group DETR~\cite{chen2022group}.

\begin{table}
  \centering
    \setlength{\tabcolsep}{5pt}
    \renewcommand{\arraystretch}{1.3}
    \footnotesize    
  \caption{\textbf{Our method, Group DETR v2, establishes
  a new SoTA on the COCO {\em test-dev} leaderboard.} }
  \begin{tabular}{c|cccccc}
    \toprule
    Method &  mAP & AP$50$ &  AP$75$ & AP$_s$ & AP$_m$ & AP$_l$ \\
    \midrule
    Group DETR v2  & $\mathbf{64.5}$ & $81.8$ & $71.1$ & $48.4$ & $67.2$ & $77.1$ \\
    \bottomrule
  \end{tabular}
  \label{tab:coco_dev}
\end{table}

\subsection{Implementation} \label{details}
The training process includes three stages: 
(i) pretrain and finetune the ViT-Huge encoder on ImageNet-1K~\cite{deng2009imagenet}, 
(ii) pretrain the whole detector (encoder and decoder) on Object365~\cite{shao2019objects365}, 
and (iii) finetune the detector on COCO~\cite{lin2014microsoft}. When pretraining the detector on Object365, 
we follow DINO~\cite{zhang2022dino} to only leave the first $5$k out of $80$k validation images as the validation set 
and add the other images to the training set. 
We also use other schemes when training the detector on Object365 and COCO, 
such as enlarging the image size to $1.5\times$ when finetuning and adopting test time augmentation. 
In addition, we apply the exponential moving average (EMA) technique~\cite{tarvainen2017mean}, use $100$ DN queries~\cite{zhang2022dino}, and adopt $11$ groups with Group DETR~\cite{chen2022group} during detector pretraining and finetuning. When finetuning the detector on COCO, we find that applying learning rate decay~\cite{clark2020electra,bao2021beit,he2022masked,chen2022context} for the components of the detector (encoder and decoder) gives a $\sim$$0.9$ mAP gain on COCO.

\section{Experiments}

\paragraph{Results on Object365 $\textbf{5}$k val.} 
We pretrain Group DETR v2 for $24$ epochs with $64$ A100 GPUs on Object365. 
On the first $5$k validation set, our Group DETR v2 achieves $\textbf{55.6}$ mAP. Table~\ref{tab:o365} gives detailed results.

\paragraph{Results on the COCO {\em test-dev}.} 
We finetune the detector
(pretrained on Object365) 
on the COCO training set
for $20$ epochs with $32$ A100 GPUs. 
During testing, we adopt test time augmentation 
with various scales and their flipped counterparts,
and perform fusion on the query features\footnote{According to our experiments, the fusion on the query features builds a robust feature across different scales and gives $\sim$$0.8$ mAP improvement.} and the final predictions~\cite{zhang2022dino}. 
Our Group DETR v2 achieves $\textbf{64.5}$ mAP on COCO {\em test-dev}. 
Table~\ref{tab:coco_dev} provides detailed results. 

\paragraph{Comparisons with state-of-the-art results on the COCO {\em test-dev} leaderboard.} 
We report the previous SoTA results on the COCO leaderboard. Table~\ref{tab:leaderboard} shows that only pretraining the ViT-Huge encoder on ImageNet-$1$K, Group DETR v2 outperforms other methods with larger models ({e.g.}, BEIT-3~\cite{wang2022image} and SwinV2-G~\cite{liu2022swin}) and more training data, and sets a new record on COCO {\em test-dev}. 
We expect that the results will be further improved with more data and larger models.

\begin{table}
  \centering
    \setlength{\tabcolsep}{5pt}
    \renewcommand{\arraystretch}{1.3}
    \footnotesize    
  \caption{\textbf{Results on Object365 $\textbf{5}$k val} with a single scale of $800\times 1333$.}
  \begin{tabular}{c|cccccc}
    \toprule
    Method &  mAP & AP$50$ &  AP$75$ & AP$_s$ & AP$_m$ & AP$_l$ \\
    \midrule
    Group DETR v2  & $\mathbf{55.6}$ & $68.8$ & $60.9$ & $36.6$ & $57.5$ & $71.3$ \\
    \bottomrule
  \end{tabular}
  \label{tab:o365}
\end{table}

{\small
\bibliographystyle{ieee_fullname}
\bibliography{egbib}
}

\end{document}